\documentclass[sigconf]{acmart}
\pdfoutput=1
\usepackage{bbold}
\usepackage{subfig}
\usepackage{graphicx}
\usepackage{float}

\newcommand{\change}[1]%
    {\textcolor{black}{#1}}

\AtBeginDocument{%
  \providecommand\BibTeX{{%
    \normalfont B\kern-0.5em{\scshape i\kern-0.25em b}\kern-0.8em\TeX}}}

\copyrightyear{2020}
\acmYear{2020}
\setcopyright{acmlicensed}
\acmConference[HRI '20]{Proceedings of the 2020 ACM/IEEE International Conference on Human-Robot Interaction}{March 23--26, 2020}{Cambridge, United Kingdom}
\acmBooktitle{Proceedings of the 2020 ACM/IEEE International Conference on Human-Robot Interaction (HRI '20), March 23--26, 2020, Cambridge, United Kingdom}
\acmPrice{15.00}
\acmDOI{10.1145/3319502.3374811}
\acmISBN{978-1-4503-6746-2/20/03}

\newcommand{\kla}{\textit{KLAggregate}}
\newcommand{\tpost}{\textit{TruePosterior}}
\newcommand{\tmatch}{\textit{TrueMatch}}
\newcommand{\ttable}{\textit{table}}
\newcommand{\thuman}{\textit{human}}
\newcommand{\tlaptop}{\textit{laptop}}
\usepackage{color}
\newcommand{\adnote}[1]{\textcolor{blue}{A: #1}}
\usepackage[normalem]{ulem}

\newcommand{\Ocal}{\mathcal{O}}
\newcommand{\Rbb}{\mathbb{R}}

\DeclareMathOperator*{\argmax}{arg\,max}

\begin{document}
\fancyhead{}

\title{LESS is More: \\Rethinking Probabilistic Models of Human Behavior}


\author{Andreea Bobu}
\authornote{
    Both authors contributed equally to this research.\\
    This research is supported by the Air Force Office of Scientific Research (AFOSR), the NSF grant IIS1734633 (SCHooL), and the NSF grant CNS1545126 (VeHICaL).
}
\affiliation{%
  \institution{University of California, Berkeley}
}
\email{abobu@berkeley.edu}

\author{Dexter R.R. Scobee}
\authornotemark[1]
\affiliation{%
  \institution{University of California, Berkeley}
}
\email{dscobee@berkeley.edu}

\author{Jaime F. Fisac}
\affiliation{%
  \institution{University of California, Berkeley}
}
\email{jfisac@berkeley.edu}

\author{S. Shankar Sastry}
\affiliation{%
  \institution{University of California, Berkeley}
}
\email{sastry@berkeley.edu}

\author{Anca D. Dragan}
\affiliation{%
 \institution{University of California, Berkeley}
}
\email{anca@berkeley.edu}

\renewcommand{\shortauthors}{Bobu and Scobee, et al.}

\begin{abstract} 
Robots need models of human behavior for both inferring human goals and preferences, and predicting what people will do. A common model is the Boltzmann noisily-rational decision model, which assumes people approximately optimize a reward function and choose trajectories in proportion to their exponentiated reward. While this model has been successful in a variety of robotics domains, its roots lie in econometrics, and in modeling decisions among different discrete options, each with its own utility or reward. In contrast, human trajectories lie in a continuous space, with continuous-valued features that influence the reward function. We propose that it is time to rethink the Boltzmann model, and design it from the ground up to operate over such trajectory spaces. 
We introduce a model that explicitly accounts for distances between trajectories, rather than only their rewards. 
Rather than each trajectory affecting the decision independently, similar trajectories now affect the decision together. 
We start by showing that our model better explains human behavior in a user study. 
We then analyze the implications this has for robot inference, first in toy environments where we have ground truth and find more accurate inference, and finally for a 7DOF robot arm learning from user demonstrations.


\end{abstract}


\maketitle
\section{Introduction}
\label{sec:intro}

\begin{figure}
  \includegraphics[width=0.898\columnwidth]{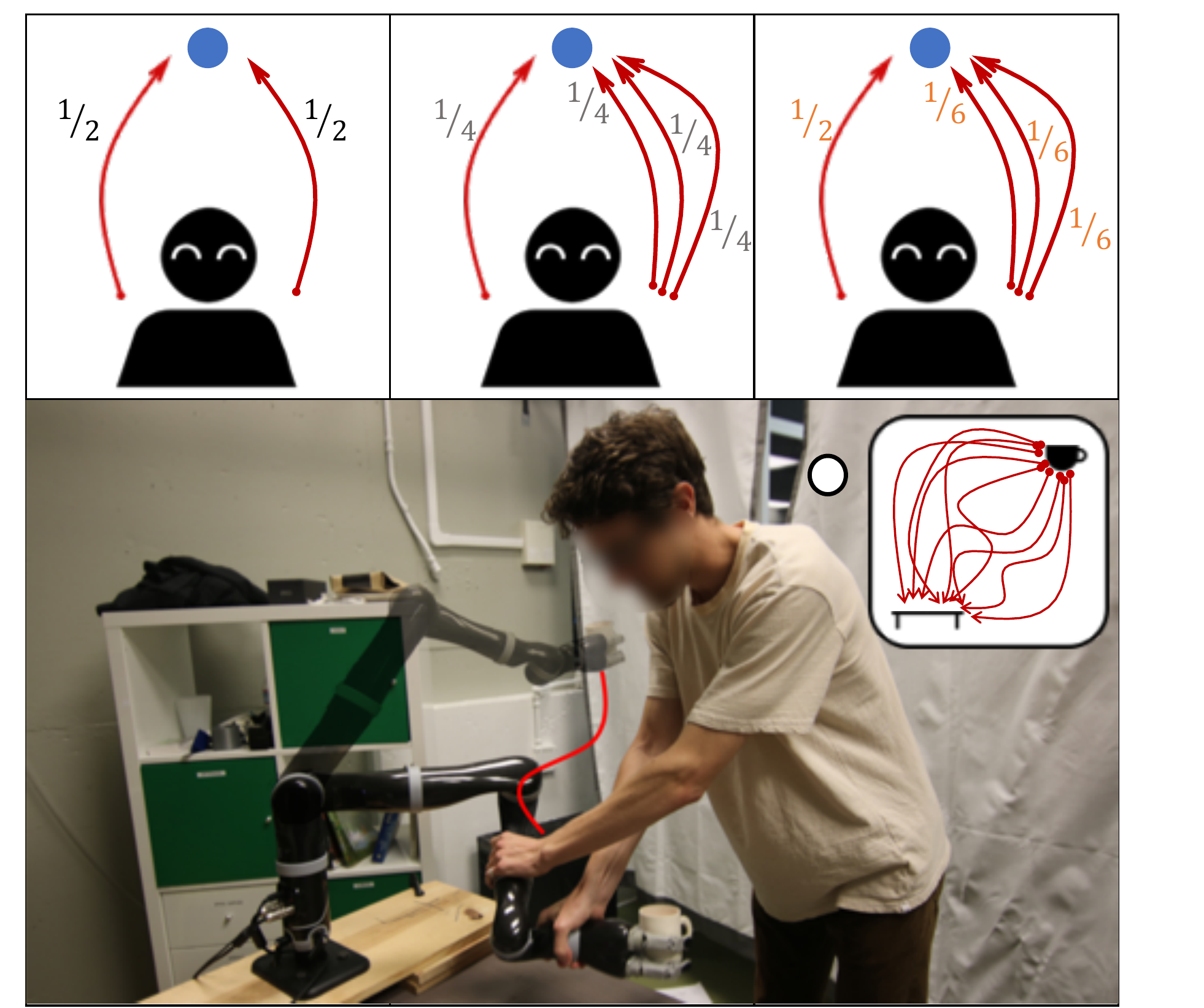}
  \caption{(Top) Contrary to Boltzmann, when adding more options to the right, LESS (right) does not drastically reduce the probability of selecting the left option. (Bottom) We test LESS on learning from user demonstrations for a 7DOF arm.}
  \label{fig:frontfig}
\end{figure}

What we do depends on our intent -- our goals and our preferences. When robots collaborate with us, they need to be able to observe our behavior and infer our intent from it, so that they can help us achieve it. They also need to anticipate or predict our future behavior given what they have inferred, so that they can seamlessly coordinate their behavior with ours. Both inference and prediction thus require a model of human behavior conditioned on intent.

A very popular such model is Boltzmann rationality~\cite{baker2007goalinference, von1945theory}. It formalizes intent via a reward function, and models the human as selecting trajectories in proportion to their (exponentiated) reward. Boltzmann rationality has seen great successes in a variety of robotic domains, from mobile robots~\cite{Kretzschmar2016socially,vasquez2014crowds,henry2010crowded,Ziebart:2009:PPP:1732643.1732694,pfeiffer2016predicting} to autonomous cars~\cite{maxent,Wulfmeier2015MaximumED,kitani2012activity} to manipulation~\cite{kalakrishnan2013learning,Bobu2018LearningUM,Finn2016gcl,mainprice2013manipulation,mainprice2015predicting}, in both inference~\cite{levine2012cioc,maxent, kalakrishnan2013learning,Kretzschmar2016socially,Finn2016gcl,Ramachandran2007birl,aghasadeghi2011pathintegrals,vasquez2014crowds,henry2010crowded} and prediction~\cite{kitani2012activity, mainprice2013manipulation,Ziebart:2009:PPP:1732643.1732694,mainprice2015predicting,pfeiffer2016predicting}.

Despite its widespread use, Boltzmann predictions are not always the most natural. At the core of the Boltzmann model is the view that behavior is a choice among available alternatives; the probability of any trajectory thus heavily depends on the available alternatives.
This has some unforeseen side-effects. One of the simplest examples is at the top of Figure \ref{fig:frontfig}. Imagine first that there are two possible trajectories to a goal, left and right, both equally good. Boltzmann would predict a $~.5$ probability of choosing to go to the left. Next, imagine that we change the set of alternatives: we add two similar trajectories to the right. Just because there are more options to go to the right, Boltzmann now predicts a higher probability that you will decide to do so: for these four equally good trajectories, Boltzmann assigns $~.25$ probability each, and estimates going left with only $.25$ probability instead of $.5$ as before. Should this change in alternatives -- the addition of similar options to go to the right -- really be reducing the prediction that you will go left by \emph{that} much?

This example seems artificial -- when are we going to have a) a group of similar trajectories, and b) an imbalance in the number of similar trajectories for each option,
so that Boltzmann shows this side-effect?
Unfortunately, it is quite representative of real-world trajectory spaces.
Spaces of trajectories are \emph{continuous and bounded},
so they naturally contain a \emph{continuum} of alternatives of varying similarity to each other, just like the right-side trajectories in our example.
Further, trajectories will have varying amounts of similarity to the rest of the space: just like our left-side trajectory
was dissimilar from the other alternatives,
in the real world, trajectories closer to joint limits or that squeeze in between two nearby obstacles will be dissimilar from the rest of the trajectory space.

Unfortunately, the Boltzmann model was not designed to handle such spaces. It has its roots in the Luce axiom of choice from econometrics and mathematical psychology \cite{Luce1977choice,Luce1959choice}, which models decisions among \emph{discrete and different} options. When we move to trajectory spaces, the options now are all connected to some degree:

\emph{Our insight is that we need to rethink how to generalize the Luce axiom to trajectory spaces, and account for how \textbf{similarity} in trajectories should influence their probability.}

We \change{take a first step towards this goal by} introducing an alternative to the Boltzmann model that accounts not just for the reward of each trajectory, but also for the feature-space similarity each trajectory has with all other alternatives.
We name our model LESS, as it is Limiting Errors due to Similar Selections.
We start by testing that our model does better at predicting human decision (Section \ref{sec:observationexp}), and then move on to analyze its implications for inference.
We first conduct experiments in simulation, with ground truth reward functions, to show that we can make more accurate inferences using our model (Section \ref{sec:inferenceexp}).
Finally, we test inference on real manipulation tasks with a 7DOF arm, where we learn from user demonstrations (Section \ref{sec:robustnessexp})-- though we no longer have ground truth, we show that we can improve the robustness of the inference if we use LESS. 
\section{Method}
\label{sec:method}




Motivated by human prediction and reward inference for robotics, we seek an improved human behavior model, explicitly designed for \emph{trajectory} spaces rather than abstract discrete decisions. To develop this theory, we first turn to the literature on human decision making.

\subsection{Background}

\subsubsection{Human Decision Making}

One of the preeminent theories of human decision making in mathematical psychology is based on Luce's axiom of choice \cite{Luce1959choice, Luce1977choice}.
In this formulation, we consider a set of options $\Ocal$, and we seek to quantify the likelihood that a human will select any particular option $o \in \Ocal$.
The desirability of each option can be modeled by a function $v: \Ocal \to \Rbb^+$, where $v$ produces higher values for more desirable options.
As a consequence of Luce's choice axiom, the probability of selecting an option $o$ is given by
\begin{equation} \label{eq:luce}
    P(o)
    = \frac{v(o)}{
        \change{\sum}_{\bar o\in\Ocal} v(\bar o)
    }
    \enspace. 
\end{equation}
If we further assume that each option $o$ has some underlying reward $R(o) \in \Rbb$, and we allow desirability to be an exponential function of this reward, then we recover the Luce-Shepard choice rule \cite{Shepard1957choice}:
\begin{equation} \label{eq:luce_shepard}
    P(o)
    = \frac{e^{R(o)}}{
        \change{\sum}_{\bar o\in\Ocal} e^{R(\bar o)}
    }
    \enspace. 
\end{equation}

When the options being chosen by the human are trajectories $\xi \in \Xi$, i.e. sequences of (potentially continuous-valued) actions, we refer to \eqref{eq:luce_shepard} as the Boltzmann model of noisily-rational behavior \cite{von1945theory,baker2007goalinference}.
The reward $R$ is typically a function of \change{a feature vector} $\phi: \Xi \to \Rbb^k$,
\change{giving the probability density $p$ over continuous $\Xi$ as}
\begin{equation} \label{eq:boltzmann}
    \change{p}(\xi)
    = \frac{e^{R(\phi(\xi))}}{
        \int_{\Xi} e^{R(\phi(\bar\xi))}d\bar\xi
    }
    \enspace. 
\end{equation}

\subsubsection{Handling duplicates}
Since the introduction of the Luce choice axiom, related works \cite{debreu1060review, Gul2014RandomCA} have pointed out its \textit{duplicates problem}, where inserting a duplicate of any option $o$ into $\Ocal$ has an undue influence on selection probabilities.
To address this drawback, various extensions of the Luce model have been proposed which attempt to group together identical or similar options \cite{benakiva1973nested, vovsha1997CNL}. 
Further extending these ideas, \citet{Gul2014RandomCA} recently introduced the \textit{attribute rule}, which reinterprets options as bundles of attributes but maintains Luce's idea that choice is governed by desirability values. 

Analogous to \cite{Gul2014RandomCA}, let $\mathcal{X}$ be the set of all attributes, let $\mathcal{X}_{o} \subseteq \mathcal{X}$ be the set of attributes belonging to $o$, and let $\mathcal{X}_{\Ocal} \subseteq \mathcal{X}$ be the set of attributes which belong to at least one option $o \in \Ocal$. Define an \textit{attribute value}, $w: \mathcal{X} \rightarrow \mathbb{R}^+$, that maps attributes to their desirability, and an \textit{attribute intensity}, $s: \mathcal{X} \times \Ocal \rightarrow \mathbb{N}$, that maps pairs of attributes and options to natural numbers, usually 0 or 1, to indicate the degree to which an attribute is expressed. For instance, an attribute could be the property ``green'' and $s(\text{``green''}, o)$ could return 1 if option $o$, say one of a set of cars, is green, and 0 otherwise.

According to the attribute rule, the probability of choosing $o$ is
\begin{equation} \label{eq:attribute}
    P(o) = \sum_{x \in \mathcal{X}_{o}}
    \frac{w(x)}{
        \sum_{\bar{x}\in\mathcal{X}_\Ocal} w(\bar{x})
    } \cdot \frac{s(x, o)}{
        \sum_{\bar{o} \in \Ocal} s(x,\bar{o})
    }
    \enspace, 
\end{equation}
%
which describes a process where the human first chooses an attribute $x\in \mathcal{X}_{\Ocal}$ according to a Luce-like rule, then an option $o \in \Ocal$ with that attribute according to another Luce-like rule. Note that \eqref{eq:attribute} reduces to \eqref{eq:luce}
if no pair of options in $\Ocal$ shares any attributes; for example, if each $o$ has a single unique attribute, the first sum in \eqref{eq:attribute} disappears, and the second fraction evaluates to 1.
In this work, we want to take advantage of the attribute rule's graceful handling of duplicates while extending its functionality to trajectories with continuous-valued features and not only categorical attributes.

\subsection{The LESS Human Decision Model}
In this paper, we take inspiration from the attribute rule to derive a novel model of human decision making in continuous spaces.
Key to our approach is introducing a similarity measure on trajectories. This could be directly in the trajectory space, but more generally it is in \emph{feature} space, where features could, in one extreme, be the trajectory itself.
We first instantiate the attribute rule with features as the attributes, and then soften it to account for feature similarity.
Indeed, the Boltzmann rationality model given by \eqref{eq:boltzmann} already assigns selection probabilities based only on trajectory features, so we look to modify the decision space to depend directly on features as well.

\subsubsection{Accounting for Trajectories with Identical Features.}

We derive our model by starting from \eqref{eq:attribute} and defining the set of attributes to be $\Phi$, the set of all possible feature vectors. Accordingly, the set of attributes that belong to $\xi$ is a single element $\Phi_\xi=\{\phi(\xi)\}$, and the attributes represented in \change{a set} $\change{\Xi' \subseteq \Xi}$ are ${\Phi_{\Xi'} = \{\phi({\xi'}) \mid \xi'\in\Xi'\}}$.
Combining this convention with the reward model \eqref{eq:boltzmann}, the modified attribute rule for trajectories \change{over a finite subset $\Xi_f \subset \Xi$} becomes 
\begin{equation} \label{eq:attribute_modified}
    P(\xi) = 
    \frac{e^{R(\phi(\xi))}}{
        \change{\sum}_{\bar\phi\in\Phi_{\Xi_f}} e^{R(\bar\phi)}
    } \cdot \frac{s(\phi(\xi),\xi)}{
        \change{\sum}_{\bar\xi\in\Xi_f} s(\phi(\xi),\bar{\xi})
    }
    \enspace. 
\end{equation}
In the original attribute rule, the attribute intensity $s$ mapped to the natural numbers.
A convenient mapping in this context would be to use $s$ as an indicator function, where $s(x,\xi)$ evaluates to 1 only if $x = \phi(\xi)$.
With this formulation, \emph{if} all trajectories have a unique feature vector, then the rightmost term of \eqref{eq:attribute_modified} is identically 1 and we recover the Boltzmann model  \eqref{eq:boltzmann}, \change{as applied to a finite sample of trajectories $\Xi_f$}.
If, on the other hand, \change{multiple} trajectories share the exact same feature vector, then they will effectively be considered as a single option, and the selection probability will be distributed equally among them.
This effect is desirable: since the features $\phi(\xi)$ capture all the relevant inputs to the reward, trajectories with the same features should be considered practically equivalent.

\subsubsection{Softening to Feature Similarity.}

We
suggest that such a
notion of attribute intensity is too stringent \change{for continuous spaces}, and we
\change{redefine}
$s$ to be a soft \textit{similarity metric} $s: \Phi \times \Xi \rightarrow \mathbb{R}^+$, which should be symmetric ($s(\phi(\xi), \bar\xi) = s(\phi(\bar\xi), \xi)$) and positive semidefinite ($s(x, \xi) \geq 0$),
with $s(\phi(\xi), \xi) = \max_{x \in \Phi, \bar\xi \in \Xi} s(x, \bar\xi)$ for all $\xi \in \Xi$. 

Using this redefined similarity metric $s$, 
\change{we extend \eqref{eq:attribute_modified} to be a probability density on the continuous trajectory space $\Xi$, as in \eqref{eq:boltzmann}:}
%
\begin{equation} \label{eq:less_rule}
    \change{p}(\xi)
     = \frac{\frac{e^{R(\phi(\xi))}}{
        \int_{\change{\Xi}} s(\phi(\xi),\bar{\xi})\ d\bar\xi}}
        {{\int_{\change{\Xi}} \frac{e^{R(\phi(\hat\xi))}}{
        \int_{\change{\Xi}} s(\phi(\hat\xi),\bar{\xi})\ d\bar\xi}}\ d\hat\xi}
        \propto \frac{e^{R(\phi(\xi))}}{
        \int_{\Xi} s(\phi(\xi),\bar{\xi})\ d\bar\xi}
    \enspace,
\end{equation}
\change{where $s(\phi(\xi),\xi)$ and the integral over $\Phi_{\Xi}$
are omitted because they are constant over $\Xi$ and cancel out during normalization.}
%


Under this new formulation, the \change{likelihood} of selecting a trajectory is inversely proportional to its feature-space similarity with other trajectories.
This de-weighting of trajectories that are similar to others is precisely the effect we seek, and
we adopt the probability given by \eqref{eq:less_rule} as our LESS model of human decision making.

\subsection{Similarity as Density}
\label{sec:similarity_metric}

The main innovation that differentiates our model from previously proposed rules is the use of a similarity metric that reweights trajectory likelihoods based on the presence of other trajectories that are nearby in feature space.
We note that the integral of this similarity over trajectories, the denominator of \eqref{eq:less_rule}, is akin to a measure of trajectory density in feature space. 
We estimate similarity as a density by selecting our similarity metric as a kernel function and performing Kernel Density Estimation (KDE).
There are many choices of kernel functions, each parametrized by some notion of bandwidth. In our experiments, we used a radial basis function, which peaks when $x = \phi(\xi)$, then exponentially decreases the farther away $x$ and $\phi(\xi)$ are from one another in feature space:
\begin{equation}
    s(x, \xi) = \change{\left( \frac{1}{\sigma\sqrt{2\pi}} \right)} \exp\left(-\frac{\|x-\phi(\xi)\|^2}{2\sigma^2}\right),
\end{equation}
where the bandwidth $\sigma$ is an important parameter that dictates, for a given feature difference between two trajectories, how much that difference affects the ultimate similarity evaluation. Higher $\sigma$ means a higher bandwidth and makes everything look more similar.


We find an optimal bandwidth $\sigma^*$ automatically by
using a finite set of samples $\Xi_f \subset \Xi$ and maximizing the sum of the log of their summed similarities, which is equivalent to maximizing their likelihood under a probability density estimate produced by KDE: 
\begin{equation}
\sigma^* = \argmax_{\sigma \in \Rbb} \sum_{\xi \in \Xi_f} \log \left( \sum_{\bar\xi \in \Xi_f} s(\phi(\xi), \bar\xi) \right)
\enspace .
\end{equation}

\subsection{Inference and Prediction with LESS}

Let \change{$\theta\in\Theta$} parametrize the reward function $R$.
To predict what the human will do given a belief $b(\theta)$, we marginalize over $\theta$:
\begin{equation}
    \change{p}(\xi)=\int_{\change{\Theta}}b(\theta)\change{p}(\xi|\theta)d\theta
    \enspace ,
\end{equation}
with $\change{p}(\xi|\theta)$ given by \eqref{eq:less_rule}. To perform inference over $\theta$ given a human trajectory, we update our belief using Bayesian inference:
\begin{equation}
b'(\theta)= \frac{ b(\theta)\change{p}(\xi|\theta)}{\int_{\change{\Theta}}b(\bar{\theta})\change{p}(\xi|\bar{\theta})d\bar\theta}
\enspace .
  \label{eq:bayes_update}  
\end{equation}
In practice, calculating the integrals in the denominators of \eqref{eq:bayes_update} \change{and \eqref{eq:less_rule}} can be intractable, so we use a discretized set of $\theta$ parameters \change{and finite trajectory sample sets in our experiments.
The specific sampling of the trajectory choice space can significantly impact inference, and we explore its implications in Section~\ref{sec:robustnessexp}.}

%
\section{LESS as a human decision model}
\label{sec:observationexp}

\begin{figure}
    \subfloat[Control trial]{\label{fig:mech_turk_control}\includegraphics[width=0.44\columnwidth]{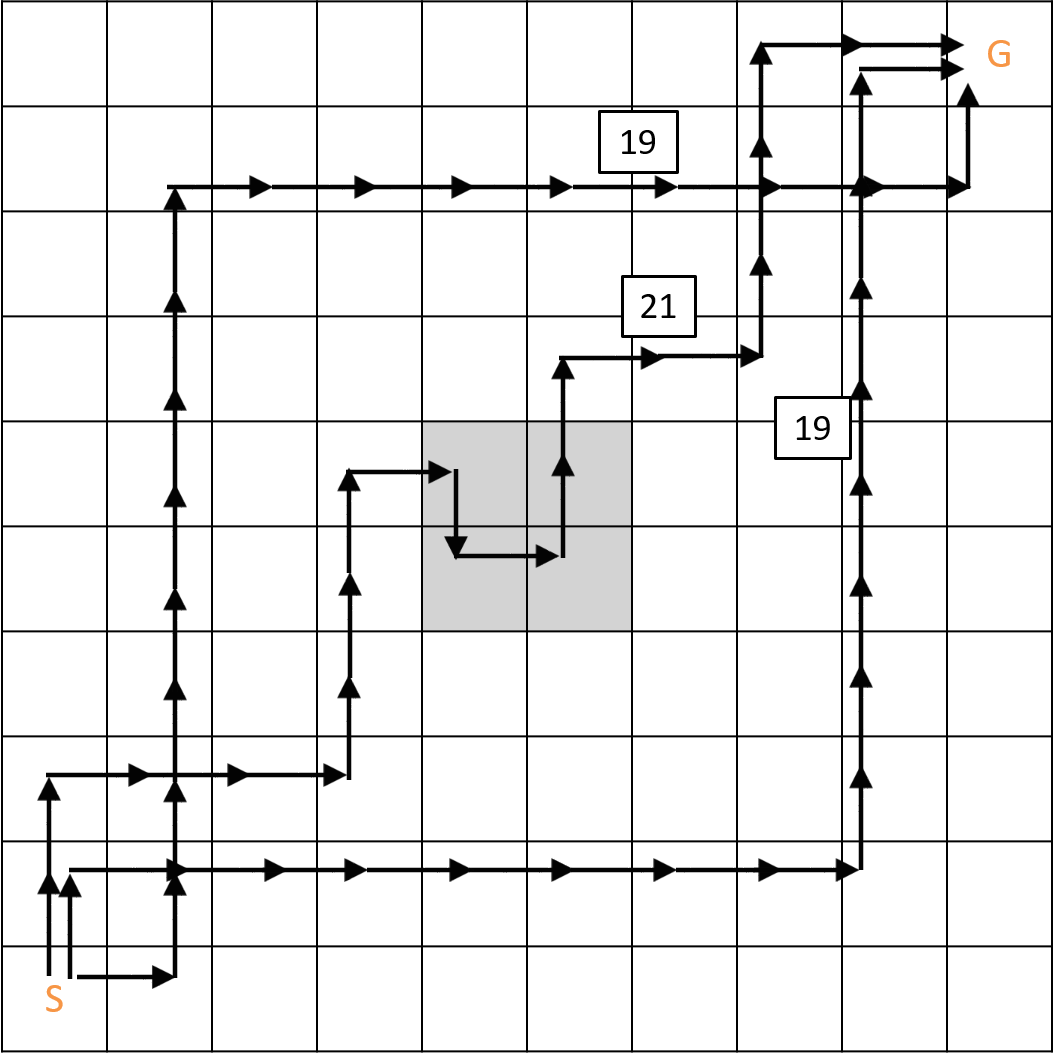}}\hfill
    \subfloat[Experimental trial]{\label{fig:mech_turk_experimental}
    \includegraphics[width=0.47\columnwidth]{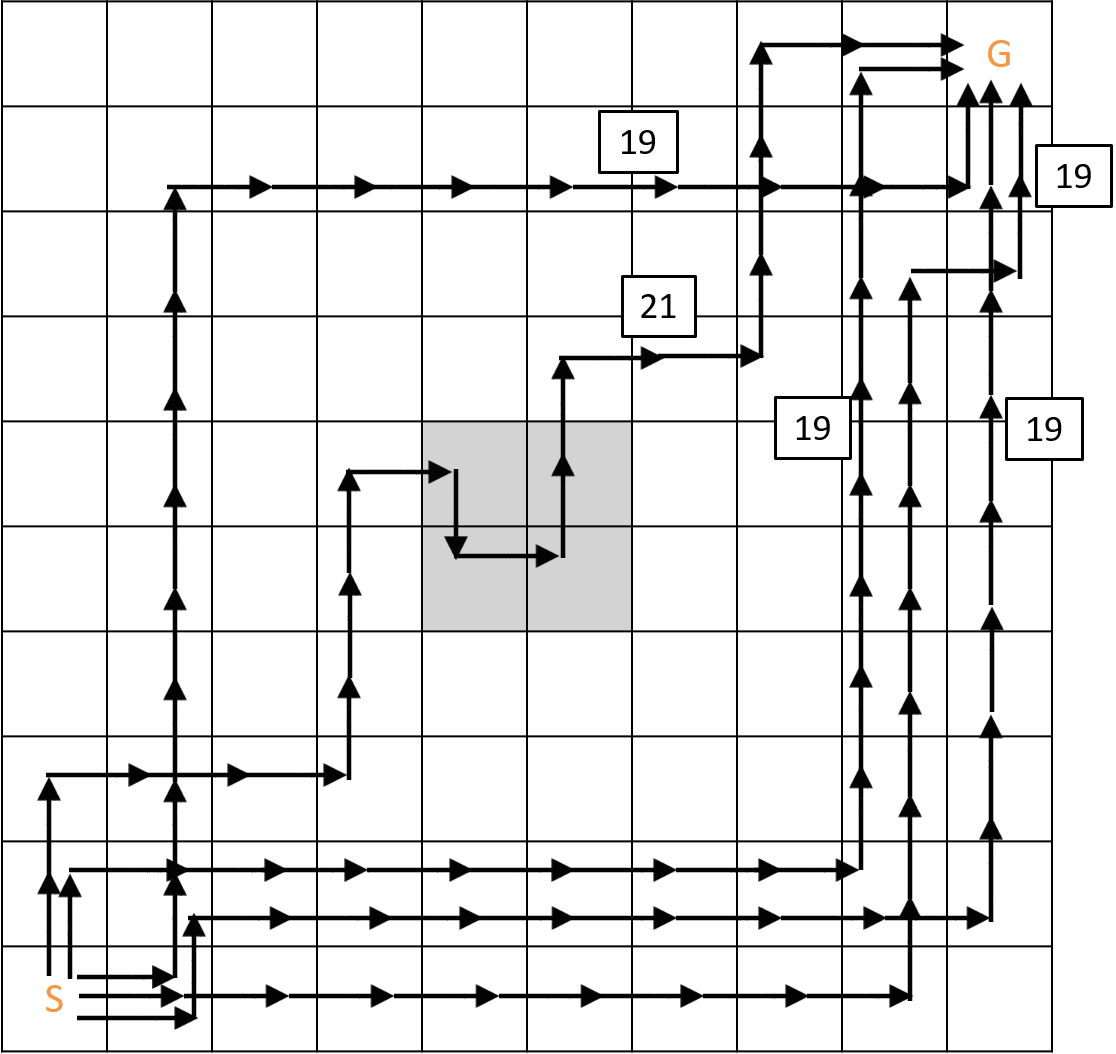}}\\
    \subfloat[Distributions for Left and Right]{\label{fig:mech_turk_left_right}\includegraphics[width=0.45\columnwidth]{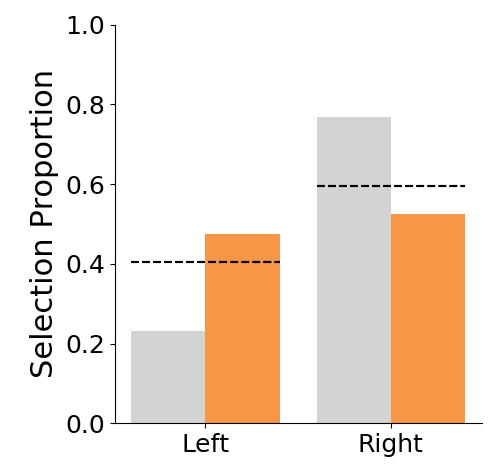}}\hfill
    \subfloat[Distributions within Right]{\label{fig:mech_turk_within_right}
    \includegraphics[width=0.45\columnwidth]{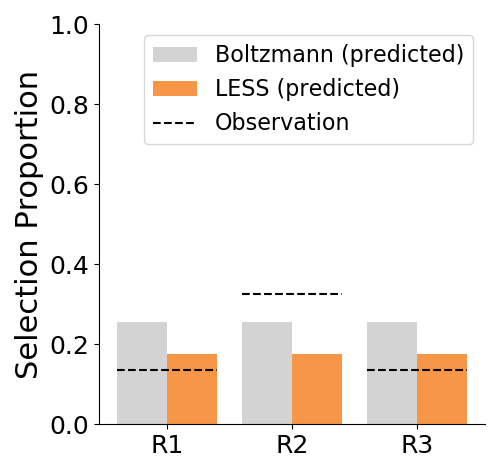}}\\
   
    \caption{The human decision model experiment.
    \protect\subref{fig:mech_turk_control} and \protect\subref{fig:mech_turk_experimental} show the trajectories used for the two trials. In \protect\subref{fig:mech_turk_left_right}, LESS predictions more closely match the observed Left-Right distribution. In \protect\subref{fig:mech_turk_within_right}, both models miss that users demonstrate a slight preference for R2 (the trajectory which visits the most states in the rightmost column in \protect\subref{fig:mech_turk_experimental}).}
    \label{fig:mech_turk}
\end{figure}

We start by testing the hypothesis that LESS is a better model for human decision making than the standard Boltzmann model.

\subsection{Human Decision Model Experiment Design}

We \change{design a browser-based user study in which we} ask participants to make behavior decisions, and measure which model best characterizes these decisions. We select a simple navigation task as our domain, where different behaviors correspond to different ways of traversing the grid from start to goal, as shown in Figure \ref{fig:mech_turk}.

\subsubsection{Main Design Idea}

The key difficulty in designing such a study is that both models require access to a ground truth reward function, i.e. user preferences over trajectories. Even though we can provide participants with some criteria -- in our case optimizing for path length while avoiding the obstacle --, this does not mean our criteria are the only ones they care about. For instance, people might implicitly prefer trajectories that go closer to or further from the obstacle, or that go around the obstacle to the left or right.

Our design idea is to introduce a control trial in which we gather data about \emph{relative} preferences among two \emph{dissimilar} options: left and right. These relative preferences then enables us to make predictions, under each model, about the experimental trial, where we add trajectories similar to the option on the right.


For the control trial, participants saw the grid world shown in Figure \ref{fig:mech_turk_control} with one obstacle in the middle and three trajectories travelling between the start and goal.
Two of the trajectories traversed an equal amount of tiles (optimal) and were symmetric along the diagonal of the grid (left and right), and a third trajectory went through the obstacle and visited more tiles than the others (not optimal).
We were only interested in what specific optimal trajectory people chose (Left versus Right), and we used the third suboptimal trajectory as an attention test to check if subjects had paid attention to the instructions.
We chose the two optimal trajectories to be symmetric and of the same color to reduce possible confounds, such as bias people might have for extraneous features like number of turns, distance from obstacle, color, etc.

For the experimental trial, shown in Figure \ref{fig:mech_turk_experimental}, we had the same setup as in the control, with the addition of two other optimal trajectories on the right. They had the same color, number of turns, and number of tiles traversed as the original right-side trajectory.
In this setup, there were two visible clusters of options: one trajectory on the left, and three clustered on the right, which we denote as the Left and Right groups, respectively.

\subsubsection{Manipulated Variables}

We manipulated the model used for decision-making in the experimental trial to be Boltzmann vs. LESS. Having access to the ratio $\lambda$ that participants chose the left trajectory over the right in the control trial means that regardless of their reward function $R(\xi)$, $e^{R(\xi_{left})}=\lambda e^{R(\xi_{right})}$, according to \eqref{eq:boltzmann}. This enables us to make predictions using both models as a function of $\lambda$ for the experimental trial, despite not knowing $R$ itself. For these computations, we assumed that all trajectories in the Right group had the same reward, that the reward of trajectories in the Left and Right groups would be equal to those estimated from the control trial, and (for LESS) that the Left trajectory had density one while the Right trajectories had density three.

Under the Boltzmann model, the addition of two trajectories similar to the one on the right decreases the probability that the trajectory on the left gets chosen. This is most obvious when $\lambda=1$, i.e. if users liked both trajectories equally -- then, $P(\xi_{\text{left}})$ would go from $.5$ all the way down to $.25$, as there are now $4$ good options. On the other hand, LESS accounts for the similarity of the trajectories on the right and keeps $P(\xi_{\text{left}})$ closer to the control value.

\subsubsection{Dependent Measures}

Our measure is the selection proportion of each trajectory in the experimental trial, which enables us to compute agreement between each model and the users' decisions.

\subsubsection{Subject Allocation}

We recruited 80 participants ($24$ female, $56$ male, with ages between $18$ and $65$) from Amazon Mechanical Turk (AMT) using the psiTurk experimental framework \cite{Gureckis2016}.
We excluded 3 participants for failing our attention test.
All participants were from the United States and had a minimum approval rating of 95\%.
The treatment trial was assigned between-subjects: participants saw only one of the sets of trajectory options.

\subsubsection{Hypotheses} \hfill

\noindent\textbf{H1:} For the experimental trial, the Boltzmann proportion prediction is significantly different from the observed proportion.

\noindent\textbf{H2:} For the experimental trial, the LESS proportion prediction is equivalent to the observed proportions.

\subsection{Analysis}
\label{sec:analysis}

In the control trial, users chose the Left trajectory 47.5\% of the time. Figure \ref{fig:mech_turk} plots the observed proportions for the experimental trial, along with each model's predictions. The experimental trial resulted in an observed probability of .41 for the Left trajectory, whereas Boltzmann predicts .23 and LESS predicts .475. The models both predict a uniform distribution among the Right trajectories. 


We performed a chi-square test of goodness of fit to see if the observed distribution of left vs. right from the experimental group differed from the predicted distributions. 
In line with our hypotheses, we found a significant difference between the observed values and the Boltzmann prediction ($X^2 (1, N = 37) = 6.27$, $p < 0.05$), and no significant difference between the observations and the LESS prediction ($X^2 (1, N = 37) = 0.72$, $p = 0.4$).

To test for equivalence, we performed an equivalence test for multinomial distributions as described by \citet{wellek2010testing}.
This test evaluates the null hypothesis that the Euclidean distance between the multinomial distribution and a reference is greater than some $\epsilon$ (where the distance is computed by taking each distribution to be a vector in $[0, 1]^k$, where $k$ is the number of trajectories represented by the distribution).
We do not have an a priori estimate for which values of $\epsilon$ are practically insignificant in this vector space of probability distributions, so we instead invert the test to find the minimum $\epsilon$ for which the observed distribution matches the predicted distribution at a significance level of $\alpha = 0.05$. We found that the minimum $\epsilon$ bound for equivalence at the $\alpha = 0.05$ level was 0.22 for the LESS prediction and 0.39 for the Boltzmann prediction.


The results across all trajectories are analogous, albeit slightly weaker because users tended to favor one of the three Right trajectories more than the other two.
The chi-square test revealed a significant difference with the Boltzmann predictions, $X^2 (1, N = 37) = 9.72$, $p < 0.05$, but no significant difference between the observations and the LESS prediction $X^2 (1, N = 37) = 5.76$, $p = 0.12$.

The equivalence test found the observed distribution matches the LESS-based predicted distribution at a significance level of $\alpha = 0.05$ when the $\epsilon$ bound is 0.29, and 0.36 for Boltzmann. \change{Despite LESS' tighter $\epsilon$, neither prediction aligns perfectly with the empirical data in Figure \ref{fig:mech_turk_within_right}.
This discrepancy is likely due to some unmodeled features (e.g. distance from the obstacle), which may influence participants' preferences.
However, while unknown features may affect both Boltzmann's and LESS' performance, LESS still corrects Boltzmann's errors from mishandling similarity.
We explore the specific effects of feature misspecification further in Section \ref{sec:misspecificationexp}.}


Overall, although neither model is a perfect predictor of behavior, we find that LESS is a better fit: Boltzmann is significantly different from the observed, and LESS provides a tighter equivalence bound.

\section{Using LESS for robot inference}
\label{sec:inferenceexp}

In Section \ref{sec:observationexp}, we provided evidence supporting that LESS can more accurately capture human decisions. This has direct implications for how robots predict behavior -- increasing the model accuracy by definition increases the robot's prediction accuracy. We now hypothesize that it also has implications for how robots \emph{infer} human preferences from behavior: namely, that using a higher accuracy model when performing inference leads to more accurate inference.

\subsection{Boltzmann and LESS inference comparison}
\label{sec:choice_by_inference}

\begin{figure}
    \subfloat[$TruePosterior$ metric for LESS sampling model.]{\label{fig:less_sampling}\includegraphics[width=\columnwidth]{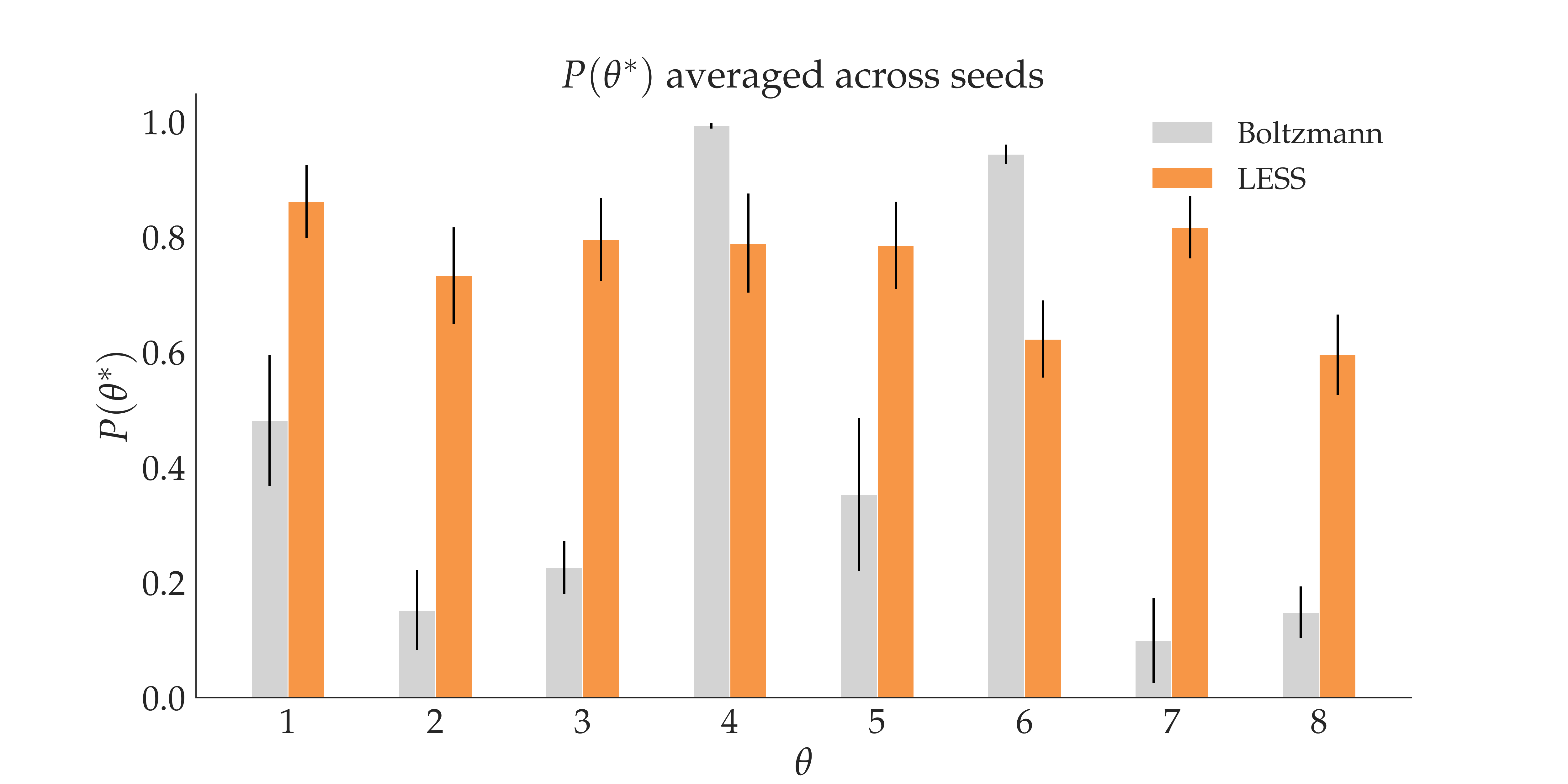}} \\
    \subfloat[$TruePosterior$ metric for Boltzmann sampling model.]{\label{fig:luce_sampling}\includegraphics[width=\columnwidth]{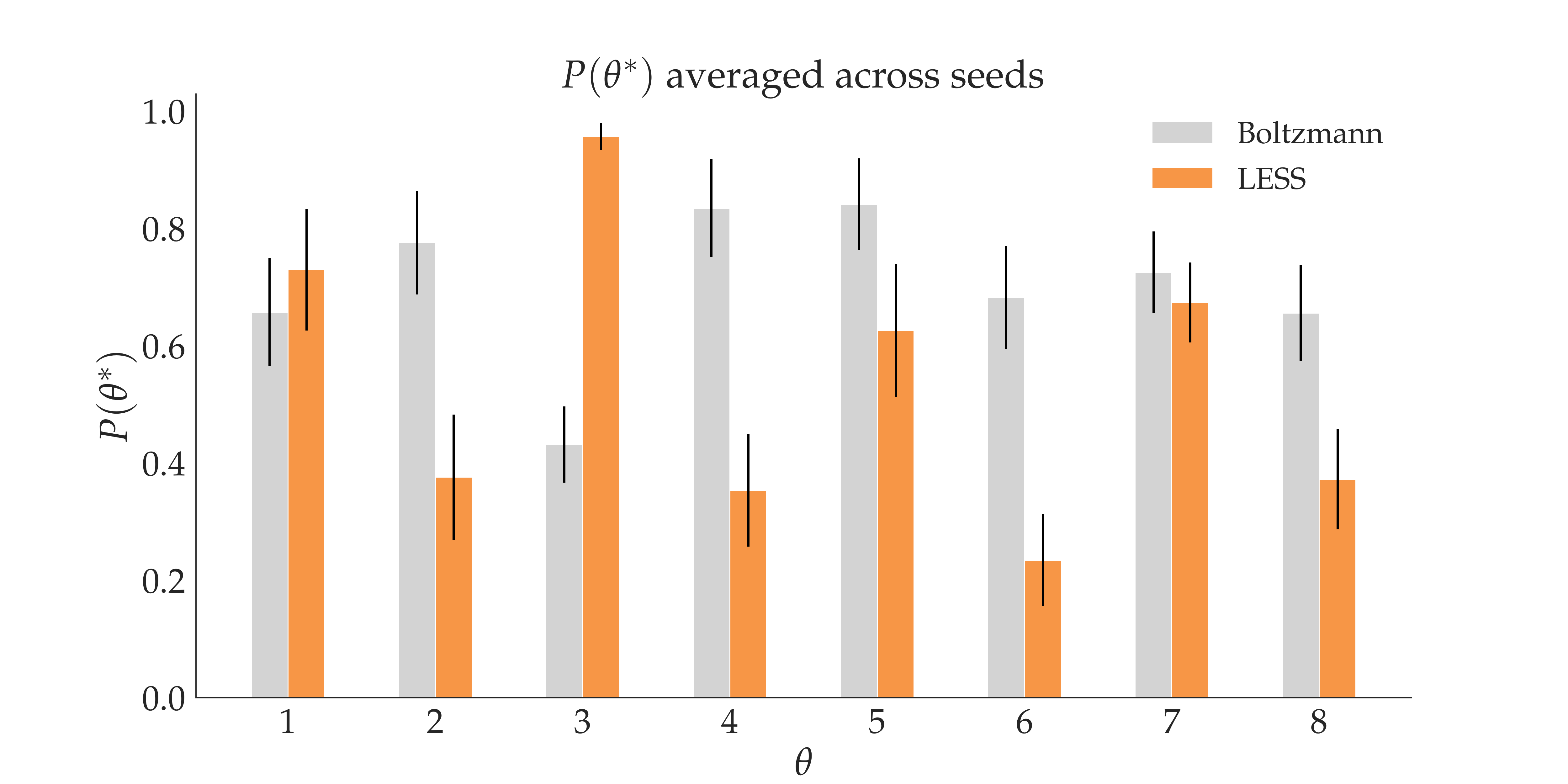}}
    \\
   
    \caption{\tpost{} results for the inference comparison experiment in Section \ref{sec:choice_by_inference}.
    Legends indicate which inference method was employed for those results.
    We found a significant interaction effect between sampling method and inference method, which can be seen in the change of relative performance for LESS and Boltzmann between \protect\subref{fig:less_sampling} and \protect\subref{fig:luce_sampling}.}
    \label{fig:inferenceaid_a}
\end{figure}

We first design an experiment to test that if people do act according to the LESS distribution, modeling them as such leads to better inference than modeling them via Boltzmann. To control for potential confounds, we also verify the opposite: if instead people acted according to Boltzmann (which Section \ref{sec:observationexp} does not support), then modeling them as Boltzmann would instead be better for inference.

In this experiment, we created a grid world environment with two objects, where humans have to teach a robot to navigate from a start to a goal and learn preferences for whether to stay close or far from the objects.
We simulated hypothetical human demonstrations $\Xi_D$ by sampling trajectories according to LESS and Boltzmann.
To do so, we fixed a particular objective $\theta^*$ and a confidence parameter $\beta$, and randomly chose trajectories according to probabilities given by either \eqref{eq:less_rule}, for LESS, or \eqref{eq:boltzmann}, for Boltzmann.
We then utilized these trajectories as ``human'' demonstrations and performed inference using either Boltzmann or LESS as the underlying choice model.
Our goal was to analyze how each model's inference quality depends on the sampling model used across a range of objectives $\theta^*$.

\subsubsection{Manipulated Variables}

We used a 2-by-2 factorial design. We manipulated the \textit{sampling model} with two levels, Boltzmann and LESS, as well as the \textit{inference model}, Boltzmann and LESS.

\subsubsection{Other Variables}

We tested inference quality across eight different $\theta^*$ values for more variation and insight.
We also used 150 random seeds for sampling demonstrations.
For a given sampling method, the combination of a $\theta^*$ and a seed determine the demonstration set that the inference will use.
Therefore, we generated 1200 demonstration sets for each sampling method.

\subsubsection{Dependent Measures}

To analyze each model's inference quality, we employ two objective metrics:\\
Accuracy of a-posteriori inference: once we obtain a posterior probability induced by the sampled $\Xi_D$, we verify that the maximum a-posteriori $\theta^{MAP}$ matches the original $\theta^*$. Thus, we define a binary variable that takes value 1 if they match and 0 otherwise:

\begin{equation*}
TrueMatch = \mathbb{1} \{\theta^{MAP} = \theta^*\}.
\end{equation*}
Magnitude of posterior $\theta^*$ probability: this metric provides a softened, continuous indication of inference performance by capturing the posterior probability mass assigned to the correct $\theta^*$:
\begin{equation*}
TruePosterior = P(\theta^* \mid \Xi_D).
\end{equation*}





\subsubsection{Hypotheses} \hfill

\noindent\textbf{H3:} When human input is generated using LESS, inference quality is significantly higher with LESS than with Boltzmann.

\noindent\textbf{H4:} When human input is generated using Boltzmann, inference quality is significantly higher with Boltzmann than with LESS.

\subsubsection{Analysis}
Figure \ref{fig:inferenceaid_a} summarizes the results by showing how \tpost{} varies by inference method for each of our sampling methods.
To analyze these results, we ran a factorial repeated measures ANOVA.
We found a significant interaction effect between the sampling and inference methods ($F(1, 1199) = 965.06$, $p < 0.001$), which can be seen with the change in relative performance of Boltzmann and LESS from Figure \ref{fig:less_sampling} to Figure \ref{fig:luce_sampling}.
A factorial logistic regression for the \tmatch{} results also revealed a significant interaction between sampling method and inference method ($p < .001$).
In post-hoc testing, a Tukey HSD test revealed that \tpost{} was significantly higher when the inference method matched the sampling method ($p < .001$ for both), and logistic regressions similarly showed that the probability of $\tmatch{} = 1$ is greater when sampling and inference agree ($p < .001$ for both).


These results strongly support both H3 and H4, as they reveal that inference performance is superior when the inference method agrees with the sampling method.
Given that the experiment in Section \ref{sec:observationexp} suggests that LESS can be a better model of human sampling behavior, these results provide evidence that using LESS-based inference could give better performance when learning from humans.

\subsection{Qualitative analysis of LESS inference}
\label{sec:qualitative}

\begin{figure}[t!]
    \subfloat[$\Xi_L$ and the resulting inferred posterior]{\label{fig:exp2qualitative}\includegraphics[width=\columnwidth]{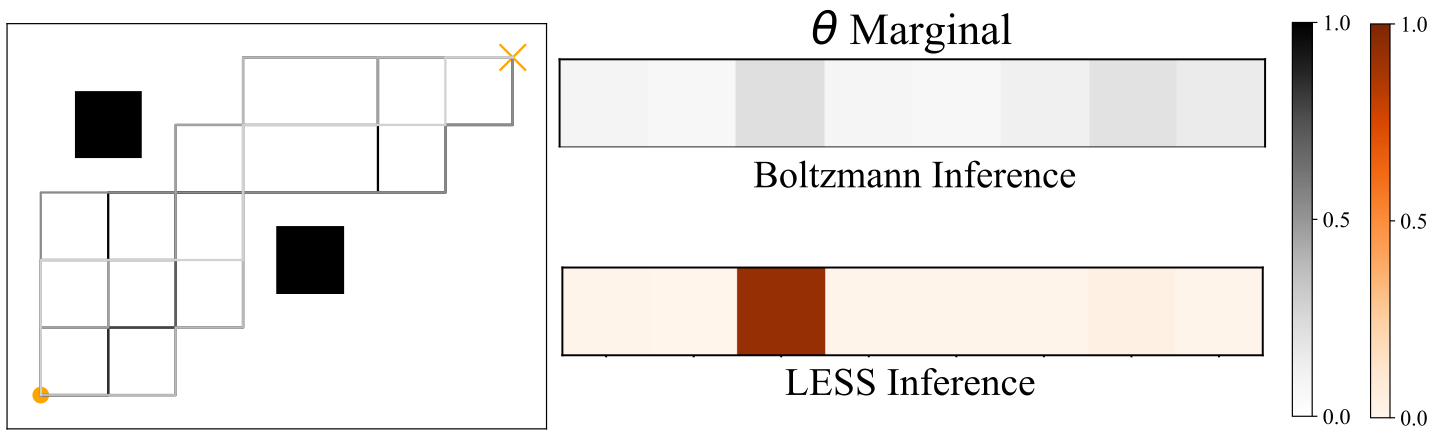}}\\
    \subfloat[$\Xi_B$ and the resulting inferred posterior]{\label{fig:exp1qualitative}\includegraphics[width=\columnwidth]{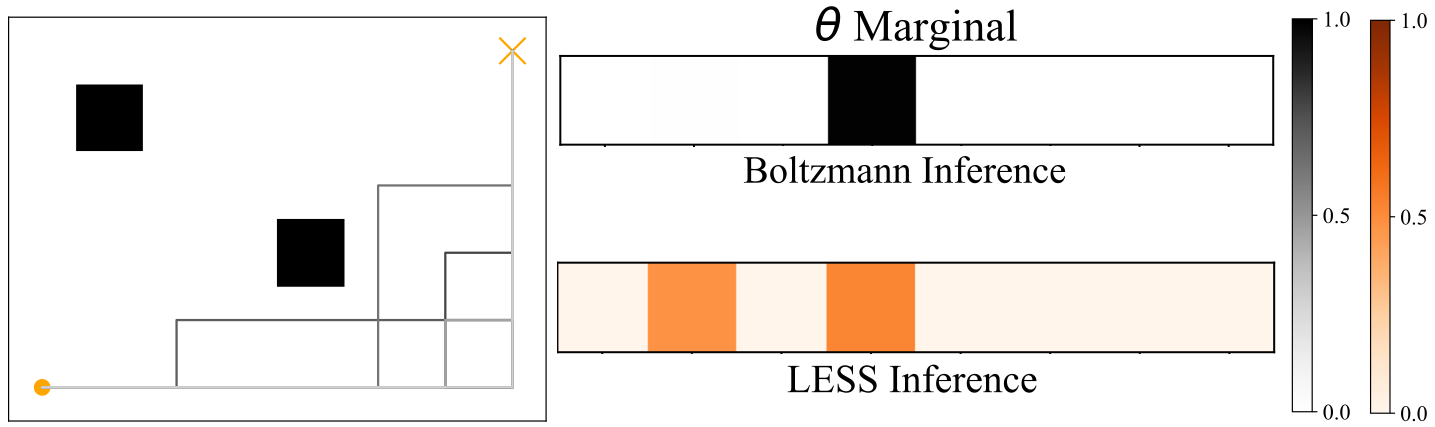}}
    \\
   
    \caption{Visualizations of $\Xi_L$ and $\Xi_B$ along with the LESS and Boltzmann inferred posteriors over $\theta$. \protect\subref{fig:exp2qualitative}: LESS learns the correct $\theta$, whereas Boltzmann under-learns. \protect\subref{fig:exp1qualitative}: Boltzmann learns the correct $\theta$, while LESS is split between avoiding both obstacles vs. avoiding the top one but being ambivalent about the bottom one.
    }
    \label{fig:qualitativegrid}
\end{figure}
\begin{figure}[t!]
    \includegraphics[width=\columnwidth]{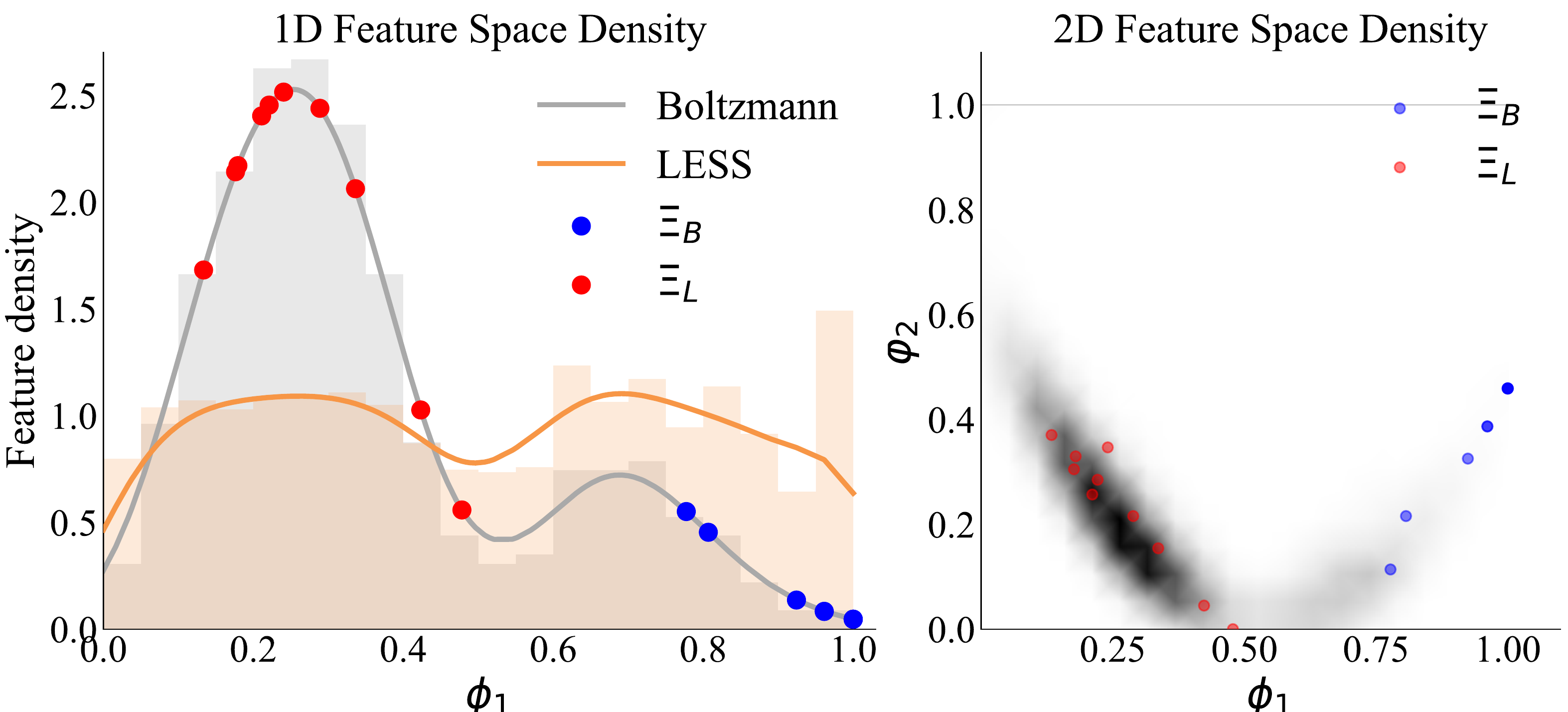}
   
    \caption{Left: actual feature density (gray), adjusted by LESS (orange). The $\Xi_L$ points (red) are in dense areas, thus Boltzmann inference under-learns. The $\Xi_B$ points are in sparse areas, but two of them are in a slightly more dense area, which makes Boltzmann reduce their relative influence and ignore the $\theta$ they suggest. Right: 2D density with $\Xi_B$, $\Xi_L$ overlaid.} 
    \label{fig:density}
\end{figure}

Based on what we have seen thus far, LESS clearly leads to different robot inferences. In this section we provide some qualitative intuition about what contributes to this difference.

The important change from Boltzmann to LESS is the \emph{strength} of the inference as a function of the feature \emph{density} at the demonstrated trajectory. If a demonstrated trajectory lies in a high-density area, i.e. its features are similar to those of many other possible trajectories, Boltzmann inference will \emph{under-learn}. This is because there are many high-reward alternatives in the normalizer of \eqref{eq:boltzmann}, which lowers the probability of the demonstration. For the analogous reason, if a demonstration lies in a low-density area, Boltzmann inference will \emph{over-learn}. Because our LESS method weighs each trajectory $\xi$ by the inverse of the density at its location in feature space $\phi(\xi)$, the resulting weighted density will be approximately uniform, not allowing the feature density to influence the strength of the inference: the presence of other options with similar features does not skew the probability as much anymore.

To visualize this, we chose two sets of demonstrations from the previous experiment. One set, $\Xi_B$, comes from one of the ground truth rewards for which Boltzmann performed better ($\theta_4$ in Figure \ref{fig:less_sampling}). The other set, $\Xi_L$, comes from one for which LESS performed better ($\theta_3$ in Figure \ref{fig:luce_sampling}).  Figure \ref{fig:qualitativegrid} shows the sampled trajectories in $\Xi_L$ and $\Xi_B$, along with the inference for each model. For $\Xi_L$, LESS confidently identifies the ground truth, whereas Boltzmann's posterior is higher entropy. Figure \ref{fig:density} shows that $\Xi_L$ does fall in a high-density region, which indeed leads to Boltzmann under-learning and finding many alternative explanations.



For $\Xi_B$, on the other hand, something very interesting happens. Looking at where the samples lie (blue dots in Figure \ref{fig:density}), two of them are in relatively high-density areas (call them $\Xi_B^{dense}$), whereas the others are in a very sparse region (call them $\Xi_B^{sparse}$). $\Xi_B^{dense}$ are the two with lower $\phi_2$ in Figure \ref{fig:density} (right). They correspond, in Figure \ref{fig:exp1qualitative}, to the two trajectories that go closer to the bottom obstacle. To the LESS inference, which is more agnostic to the feature density, this gives evidence for two hypotheses: $\Xi_B^{dense}$ support the hypothesis that the robot should stay far from the top obstacle, but be ambivalent about the bottom one, whereas the other trajectories, $\Xi_B^{sparse}$, support that the robot should stay far from both obstacles. This is why we see two hypotheses inferred by LESS in \ref{fig:exp1qualitative}. The Boltzmann inference, however, learns much more from the trajectories that lie in the low-density area, essentially ignoring $\Xi_B^{dense}$. This is what leads to the very confident inference of only one of the hypotheses. In this case, this happens to be the correct hypothesis. In general though, the opposite could have happened -- had the two trajectories that go closer to the obstacle been the ones to lie in a sparse area, Boltzmann would have confidently inferred the wrong objective. In summary, Boltzmann, by being sensitive to feature densities, can under- or over-learn.

\subsection{LESS and feature misspecification}
\label{sec:misspecificationexp}

LESS uses information from features to compute similarity, even when those features do not affect the reward. For example, if the reward is solely about efficiency, LESS captures that people treat "right-of-the-obstacle" options as similar. What if the robot does not have access though to these additional features?


\subsubsection{Experimental Design}

We again generate demonstrations using LESS, but we include two additional features: the average $x$ and average $y$ coordinate of the trajectory.  The two new features do not influence the trajectories' reward values, but they do influence the similarity metric.
To induce a misspecification, the robot performing inference is unaware of these new features.
For this experiment, we only manipulate the \textit{inference model}: LESS vs. Boltzmann.

\noindent\textbf{H5:} When the robot's feature space is misspecified, inference quality with LESS is still superior to inference quality with Boltzmann for LESS-sampled demonstrations.

\subsubsection{Analysis}
For \tpost{}, we performed a one-way repeated measures ANOVA, and as hypothesized, the test revealed that LESS inference was still significantly better than Boltzmann, in spite of the feature misspecification ($p < .001$).
Similarly for \tmatch{}, a logistic regression revealed that
the odds of having $\tmatch{} = 1$ were significantly greater when using LESS ($p < .001$), strongly supporting our hypothesis.



We take this result with a grain of salt: in the worst case, if an unspecified feature completely differentiates all options for the human, then even a human sampling according to LESS would exhibit behavior approaching the Boltzmann distribution. Then, based on Section \ref{sec:choice_by_inference}, Boltzmann inference could yield superior results.
However, this experiment suggest that in practical rather than adversarial cases, it is still preferable to use LESS inference on an incomplete set of features. Further, it is always possible to default in LESS to using the trajectory space directly for the similarity metric $s$ and not rely on features.

\section{Robust Inference for High-DOF Arms}
\label{sec:robustnessexp}

\begin{figure}
    \subfloat[\kla{} metric for single inference comparison.]{\label{fig:single_robustness}\includegraphics[width=.8\columnwidth]{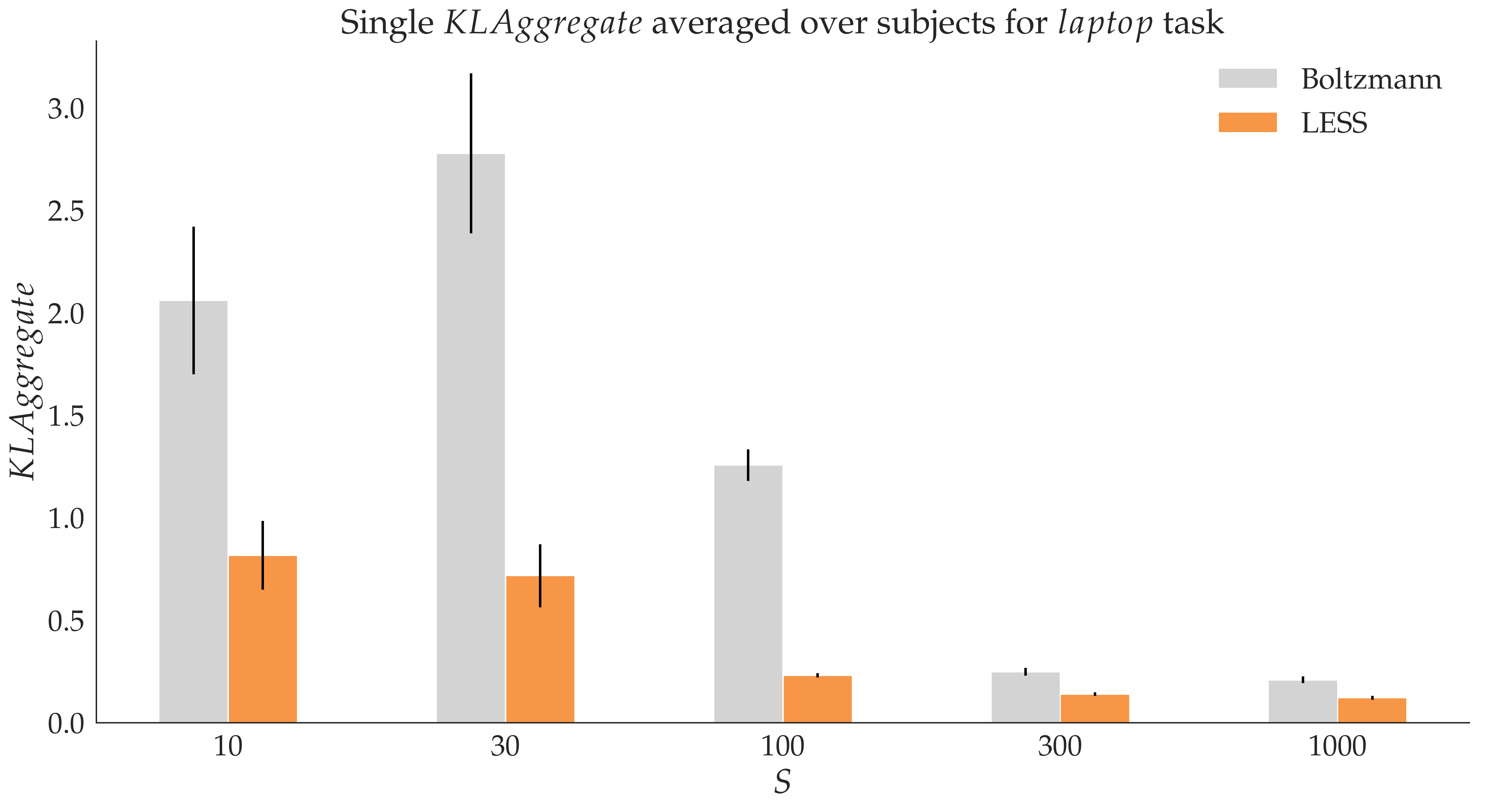}} \\
    \subfloat[$\log(\kla{})$ metric for batch inference comparison.]{\label{fig:batch_robustness}\includegraphics[width=.8\columnwidth]{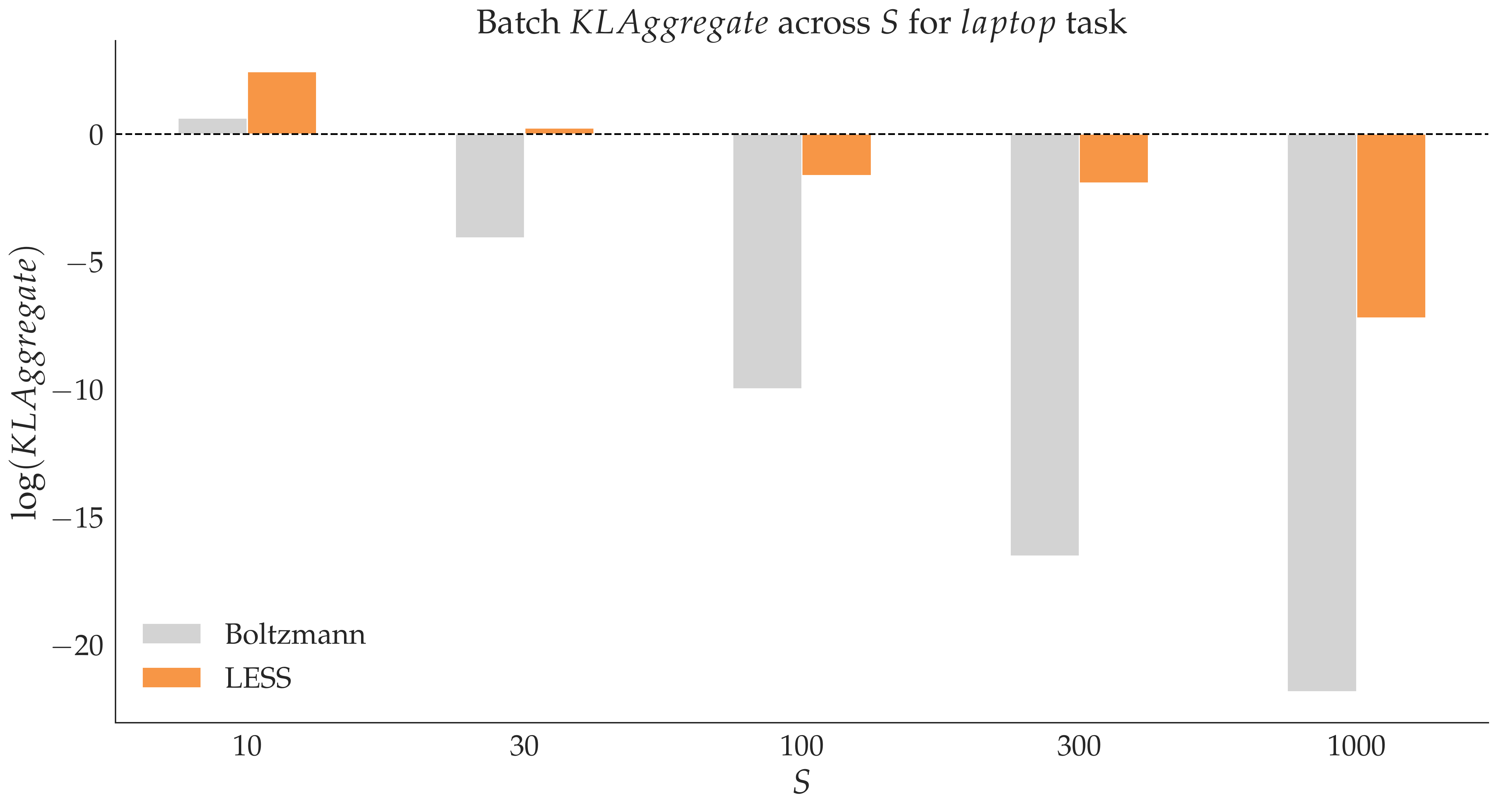}}
    \\
   
    \caption{Results for the \tlaptop{} task in the robustness analysis experiments.
    In \protect\subref{fig:single_robustness}, LESS significantly outperforms Boltzmann at low sample sizes, but they converge for the largest sample sizes.
    For the batch inference task in \protect\subref{fig:batch_robustness}, Boltzmann outperforms LESS at the lowest sample size, but the two methods converge towards zero as sample size increase.
    }
    \label{fig:robustness_a}
\end{figure}
\begin{figure}[t]
    \centering
    \includegraphics[width=\columnwidth]{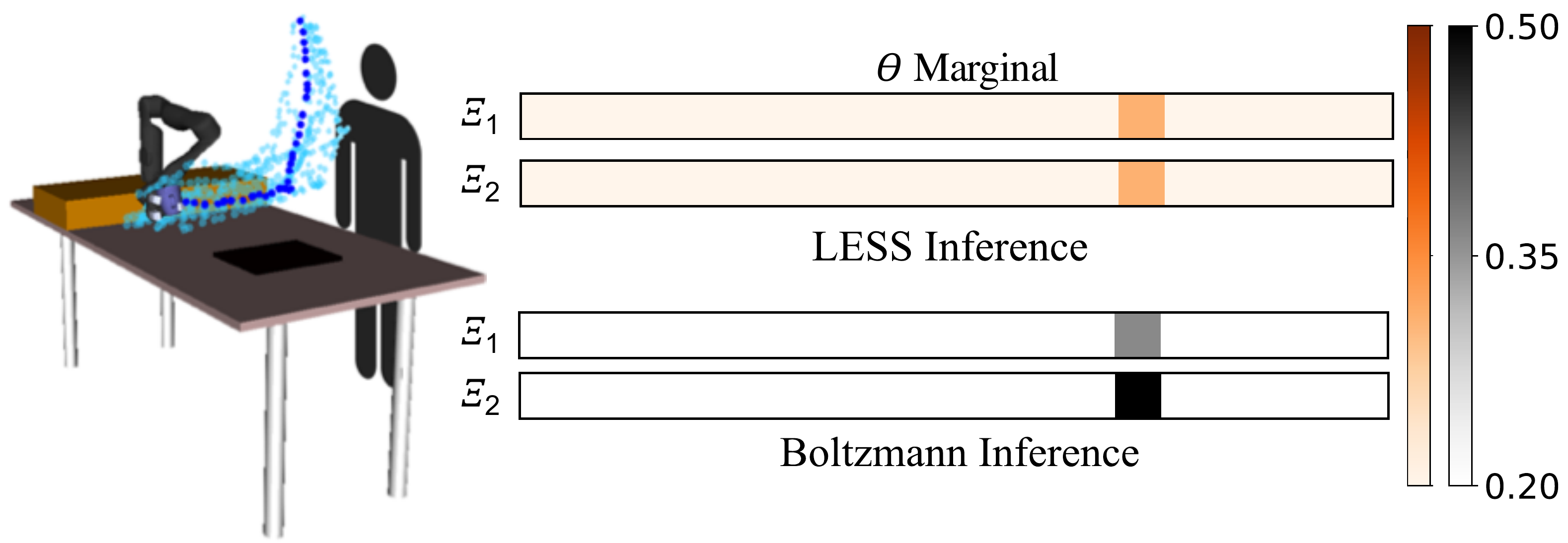}
    \caption{
    Single-demonstration (blue) inference posteriors for the \ttable{} task with two different trajectory sets of 100 samples.
    The distributions reveal that both Boltzmann and LESS produce the same $\theta^{\text{MAP}}$, but there is less variability between the LESS posteriors, leading to lower \kla{}.
    }
    \label{fig:robustnesscompare}
\end{figure}

Section \ref{sec:inferenceexp} teased that Boltzmann inference performance is highly dependent on the structure of the environment, and, more precisely, the feature space density induced by all possible trajectories. However, we demonstrated this on a toy task with simulated human data and ground truth access. We now put the same hypothesis to test in a real world high-dimensional scenario with a 7DoF robotic manipulator and real human demonstrations, where one cannot have access to the full trajectory space, nor the ground truth reward.

\subsection{Single demonstration inference}
\label{sec:single_inference_experiment}

\subsubsection{Study Goal.} Since for such an environment calculating the denominator in \eqref{eq:boltzmann} exactly is intractable, practitioners typically \emph{sample} the space of trajectories, obtaining varying subsets. Given the Boltzmann model's high dependency on the feature space density, we speculate that different sample sets would result in vastly varying inference results. In this section, we investigate how LESS can mitigate this effect and help inference robustness. We collect demonstrations from participants for different tasks, and run inference using different sets of trajectory for computing the normalizer.


\subsubsection{Manipulated Variables}
We used a 2-by-5 factorial design. We manipulated the \textit{inference model} with two levels, Boltzmann and LESS, as well as the
size $S$ of the sampled trajectory sets used for inference, with five levels: 10, 30, 100, 300, and 1000. We sample 10 different trajectory sets of each size.

\subsubsection{Other Variables}
We tested our hypothesis across three household manipulation tasks where the robot learned to carry a coffee mug from a start position to a goal according to the person's preferences.
In the first task, which we dub \ttable{}, the participants were asked to move the robot arm from start to goal while maintaining the end-effector close to the table, to prevent the mug from breaking in case of a slip. In the second task, dubbed \tlaptop{}, the participants were instructed to avoid spilling the coffee over a laptop by providing a demonstration that keeps the robot's end-effector away from the electronic device. Lastly, in the third task, dubbed \thuman{} we asked the participants to keep the end-effector away from their body, to avoid spilling coffee on their clothes.

In all scenarios, the robot performs inference by reasoning over three features: one feature of interest (distance from the table, distance from the laptop, and distance from the human, respectively), a second feature drawn from that set, and an efficiency feature computed as the sum of squared velocities across the trajectory.

\subsubsection{Dependent Measures}
In total, for each task $T$, sample size $S$, inference method $M$, and user $i$, we obtained 10 posterior distributions $P_{M,S}^{T,i}(\hat{\theta} \mid \xi^{T,i})$ constituting a set $\mathcal{P}_{M,S}^{T,i}$. Our goal was to test how robust (or consistent) each method's inference result was across the ten different trajectory sets.
We used an aggregate Kullback-Leibler divergence as a measure of how much the posterior distributions $P \in \mathcal{P}_{M,S}^{T,i}$ differ from one another:
$$KLAggregate = -\sum_{P \in \mathcal{P}_{M,S}^{T,i}}\sum_{Q \in \mathcal{P}_{M,S}^{T,i}}\sum_{\hat{\theta} \in \Theta} P(\hat{\theta} \mid \xi^{T,i}) \log \left(\frac{Q(\hat{\theta} \mid \xi^{T,i})}{P(\hat{\theta} \mid \xi^{T,i})} \right).$$
\subsubsection{Hypothesis} \hfill

\noindent\textbf{H6:} Performing single inference with LESS across multiple trajectory sets results in higher robustness and, thus, a lower \kla{} measure than inference with Boltzmann.

\subsubsection{Subject Allocation}

We recruited 12 users (3 female, 9 male, aged 18-30) from the campus community to physically interact with a JACO 7DOF robotic arm and provide demonstrations for three tasks. Figure \ref{fig:robustnesscompare} (left) illustrates the demonstrations collected for the \ttable{} task. Before giving any demonstrations, each person was allowed a period of training with the robot in gravity compensation mode, in order to get accustomed to interacting with the robot. 

\subsubsection{Analysis}
\label{sec:robot_individual_results}

As seen in Figure \ref{fig:robustnesscompare}, given two different trajectory sets, inference with each method can have drastically different outcomes. With LESS (top), we see that the resulting posterior distributions are fairly similar, whereas with Boltzmann inference (bottom), they differ in entropy/confidence.

For each sample task $T$, we performed a factorial repeated-measures ANOVA.
The results for the \tlaptop{} task are summarized in Figure \ref{fig:single_robustness}.
As the trend in the figure indicates, we found a significant interaction effect between inference method and sample size ($F(4, 44) = 40.37$, $p < .001$).
A post-hoc Tukey HSD test revealed that LESS produced significantly lower \textit{KLAggregate} than Boltzmann for $S = $ 10, 30, and 100 ($p < 0.001$ for all), but there was no significant difference found for $S = $ 300 or 1000 ($p \approx 1.00$ for both).
This trend supports our hypothesis that LESS provides more robust single-demonstration inference, and it reveals that the difference in \kla{} between LESS and Boltzmann disappears with increasing sample size. 
Results from the \ttable{} task also support this trend, with a significant main effect of inference method. 




While the \thuman{} task did reveal a significant interaction between inference method and sample size ($F(4, 44) = 2.85$, $p < .05$) it stands apart from the other two: 
a post-hoc Tukey HSD test only found a difference for sample size 1000 ($p < .001$).
This pattern indicates that demonstrations from this task may be generally more ambiguous and present a more difficult inference problem than the other two.

\subsection{All demonstrations inference}

We repeated the same experiment, except this time we run inference by aggregating all users' demonstrations for a task (batch inference). This would happen in practice if we were interested in teaching the robot about what the average user wants, rather than focusing on customizing the behavior to each user. Here, we found the opposite results, also shown in in Figure \ref{fig:batch_robustness}: LESS has higher divergence (lower robustness). We attribute this to the phenomenon described in Section  \ref{sec:qualitative}. When we had only one demonstration before, Boltzmann was not robust because, depending on the set of samples, the demonstration could fall in low- \emph{or} high-density regions, thus leading to different Boltzmann inferences for different sets. Now, with 12 demonstrations at once, the chances of one demonstration falling in a low-density area are much higher. As we've seen in Section \ref{sec:qualitative}, when there are multiple demonstrations, Boltzmann inference will be dominated by those lying in low-density areas. This leads to a more consistent posterior distribution, so long as the low-density demonstrations suggest the same reward function.

\section{Discussion}
\label{sec:conclusion}

We propose a new probabilistic human behavior model and present compelling evidence that it better captures human decision making and it attenuates inference errors that arise due to similar selections, increasing accuracy and robustness.

One limitation of our method is its reliance on a pre-specified set of robot features for similarity selection, which makes feature misspecification a possible limitation. Although our experiments in Section \ref{sec:misspecificationexp} reveal that LESS still performs better inference than Boltzmann, it is unclear whether this outcome is due to the effect of hypothesis H3 or if
our method is truly unaffected by misspecification. Further experiments are needed for complete clarification.

Our 12-person aggregate inference results in Section \ref{sec:robustnessexp} show that LESS can lead to less robust inference. We attributed this outcome to the phenomenon in Section \ref{sec:qualitative}, but it remains unclear whether this leads to less accurate inference, or whether Boltzmann is actually preferable in situations with enough varied demonstrations. 

Lastly, the Mechanical Turk study in Section \ref{sec:observationexp}, although compelling, illustrates simplistic datasets of human choices. Further studies on human behavior in more realistic settings would be useful, but complicated by lack of access to the "ground truth" reward. 

Despite these limitations, Boltzmann rationality has become so fundamental to how robots do inference and prediction, that designing a counterpart for continuous robotics domains is sorely needed. We are excited to have taken a step in this direction.



\bibliographystyle{ACM-Reference-Format}
\balance
\bibliography{hri2020}



\end{document}